\documentclass[journal]{IEEEtran}
\usepackage[pdftex]{graphicx}
\ifCLASSINFOpdf
\else
\fi
\usepackage{adjustbox}
\usepackage{booktabs}
\usepackage{xcolor}
\usepackage{amssymb}
\usepackage{multirow}
\usepackage{hyperref}
\usepackage{amsmath}

\begin{document}
%
\title{MambaOcc: Visual State Space Model for BEV-based Occupancy Prediction with Local Adaptive Reordering}
%
%
%

\author{Yonglin~Tian, Songlin~Bai, Zhiyao~Luo, Yutong~Wang, Yisheng~Lv, and Fei-Yue~Wang
\thanks{Yonglin Tian, Zhiyao Luo, Yutong Wang, Yisheng Lv and Fei-Yue Wang are with the State Key Laboratory of Multimodal Artificial Intelligence Systems, Institute of Automation, Chinese Academy of Sciences, Beijing 100190, China.}
\thanks{Songlin Bai is with the Meituan Group, Beijing 100102, China.}
}

\maketitle

\begin{abstract}
Occupancy prediction has attracted intensive attention and shown great superiority in the development of autonomous driving systems. The fine-grained environmental representation brought by occupancy prediction in terms of both geometry and semantic information has facilitated the general perception and safe planning under open scenarios. However, it also brings high computation costs and heavy parameters in existing works that utilize voxel-based 3d dense representation and Transformer-based quadratic attention. To address these challenges, in this paper, we propose a Mamba-based occupancy prediction method (MambaOcc) adopting BEV features to ease the burden of 3D scenario representation, and linear Mamba-style attention to achieve efficient long-range perception. Besides, to address the sensitivity of Mamba to sequence order, we propose a local adaptive reordering (LAR) mechanism with deformable convolution and design a hybrid BEV encoder comprised of convolution layers and Mamba. Extensive experiments on the Occ3D-nuScenes dataset demonstrate that MambaOcc achieves state-of-the-art performance in terms of both accuracy and computational efficiency. For example, compared to FlashOcc, MambaOcc delivers superior results while reducing the number of parameters by 42\% and computational costs by 39\%. Code will be available at \href{https://github.com/Hub-Tian/MambaOcc}{https://github.com/Hub-Tian/MambaOcc}.
\end{abstract}


%
\IEEEpeerreviewmaketitle

\section{Introduction}
%
%
%
%
\IEEEPARstart{P}{recise}  and stable perception of surrounding environments is crucial for autonomous vehicles (AVs). However, this becomes especially challenging when AVs navigate open scenarios, where they must detect unseen or irregular objects. Recently, occupancy prediction has garnered significant attention in the perception and planning pipelines of autonomous driving systems, owing to its fine-grained general perception capabilities.

Occupancy prediction involves determining the state of every voxel in a 3D scenario, which results in high computational and memory demands during the development of perception models. To address this issue, approaches such as multi-view planes \cite{huang2023tri} and sparse prediction methods \cite{ming2024inversematrixvt3d} have been proposed, significantly improving the efficiency of occupancy prediction. Nonetheless, these methods still fall short of efficiency compared to 2D tasks. FlashOcc \cite{yu2023flashocc} proposes leveraging BEV features to represent the entire scene and incorporates height information into the channel dimension. While it demonstrates promising results for efficient occupancy prediction, its performance heavily depends on powerful 2D backbones like Transformers \cite{liu2024swin}.

In this paper, we focus on enhancing the performance of BEV-based occupancy prediction while simultaneously reducing both parameter count and computational costs. Previous work has highlighted the advantages of Transformers in long-range modeling \cite{ma2024cotr}, but their computational burden is considerable due to quadratic attention mechanisms. Recently, state space models (SSMs), such as Mamba \cite{gu2023mamba,liu2024vmambavisualstatespace}, have emerged as more efficient solutions for long-range modeling. This development motivates us to explore the potential of state space models for improving occupancy prediction tasks.

To this end, we propose MambaOcc, a novel occupancy prediction method based on the Mamba framework, designed to be lightweight while providing efficient long-range modeling. First, we utilize the four-direction visual Mamba \cite{liu2024vmambavisualstatespace} for extracting image features. To mitigate the high computational burden associated with 3D voxels, we use BEV features as the intermediate representation for occupancy prediction and develop a hybrid BEV encoder combining convolutional layers and Mamba layers. Given the Mamba architecture's sensitivity to token ordering during feature extraction, we introduce a local adaptive reordering module that leverages deformable convolution (DCN) layers. This module is designed to dynamically udpate the context for each position, allowing the model to better capture and utilize local dependencies in the data. This approach not only mitigates the issues associated with rigid token sequences but also improves the overall accuracy of occupancy predictions by ensuring that relevant contextual information is prioritized during the extraction process. The contributions of this paper are as follows:

\begin{itemize}
    \item A lightweight Mamba-based occupancy prediction method (MambaOcc) is proposed, which boosts the performance of the BEV-based approach with significantly less computation cost. As far as we know, it is the first work that integrates Mamba into the BEV-based occupancy network. 
    \item A novel LAR-SS2D hybrid encoder with the local adaptive reordering mechanism is proposed, which enables more flexible sequence order optimization and improves the performance of state space models.
    \item We achieve state-of-the-art performance on Occ3D-nuScenes dataset with limited parameters and computations, eg, we achieve better results than FlashOcc while reducing the number of parameters by 42\% and computational costs by 39\%. 
\end{itemize}

\section{Related Works}
\subsection{3D Occupancy Prediction.}
3D occupancy prediction is vital for autonomous vehicles as it enables a comprehensive understanding of the surrounding environment by determining the occupancy status and assigning semantic labels to every voxel within a 3D traffic scene. MonoScene \cite{cao2022monoscene} was a groundbreaking work in this domain, introducing the first camera-based occupancy network capable of interpreting traffic scenes using only a monocular camera. To counteract the typical loss of vertical information associated with Bird's-Eye-View (BEV) representations, TPVFormer \cite{huang2023tri} introduced a tri-perspective view approach. OccFormer \cite{zhang2023occformer} proposed a dual-path transformer network that divides the intensive 3D processing into separate local and global transformer pathways for occupancy prediction. Moreover, OpenOccupancy \cite{wang2023openoccupancy}, Occ3D \cite{tian2024occ3d}, and SurroundOcc \cite{wei2023surroundocc} developed methods to produce high-quality dense occupancy labels. FlashOCC \cite{yu2023flashocc} introduced a BEV-based 3D representation method for fast and memory-efficient occupancy prediction while maintaining high precision. While the techniques in \cite{cao2022monoscene, huang2023tri, tian2024occ3d, wang2023openoccupancy, wei2023surroundocc, yu2023flashocc} employ CNNs for feature extraction in both image and 3D domains, CNNs have limitations in capturing global context. OccFormer \cite{zhang2023occformer} employs a dual-path transformer encoder to handle long-range features, but the quadratic complexity of transformer blocks makes their deployment in real-world scenarios challenging.

\subsection{State Space Models.}

The State Space Model (SSM) \cite{gu2021efficiently,gu2021combining,gupta2022diagonal,li2022makes,orvieto2023resurrecting} represents a class of architectures that have been adapted into deep learning as state space transforms. While SSMs offer the potential to handle tokens with long-range dependencies, they are notoriously difficult to train due to their high computational and memory demands. In response to these challenges, significant efforts \cite{gu2023mamba,gupta2022diagonal,smith2022simplified} have been made to make SSMs more competitive with mainstream architectures like CNNs and Transformers. For instance, S4 \cite{gu2021efficiently} introduces parameter normalization within a diagonal structure, while Mamba \cite{gu2023mamba} advances the design by incorporating input-specific parameterization and a scalable, hardware-optimized algorithm, resulting in a more streamlined and efficient SSM. Following the success of Mamba in text processing, researchers have begun exploring its application in computer vision. SS4ND \cite{nguyen2022s4nd} apply SSM to visual tasks, showcasing its ability to model large-scale visual data in 1D, 2D, and 3D as continuous multidimensional signals. Vmamba \cite{liu2024vmambavisualstatespace} introduces a Mamba-based vision backbone alongside a cross-scan module to address direction-sensitivity issues in images, while Vim \cite{zhu2024vision} proposes bidirectional Mamba blocks to create a flexible and efficient vision backbone. The impressive performance of the Mamba backbone in vision tasks has led researchers \cite{ma2024u,liu2024swin} to apply it in specialized visual domains, with U-Mamba \cite{ma2024u} and Swin-UMamba \cite{liu2024swin} pioneering its use in medical image segmentation. Despite these advancements, the application of Mamba-based backbones in 3D occupancy prediction remains unexplored.

\section{Method}

\begin{figure*}[h]
\centering
\includegraphics[width=7in]{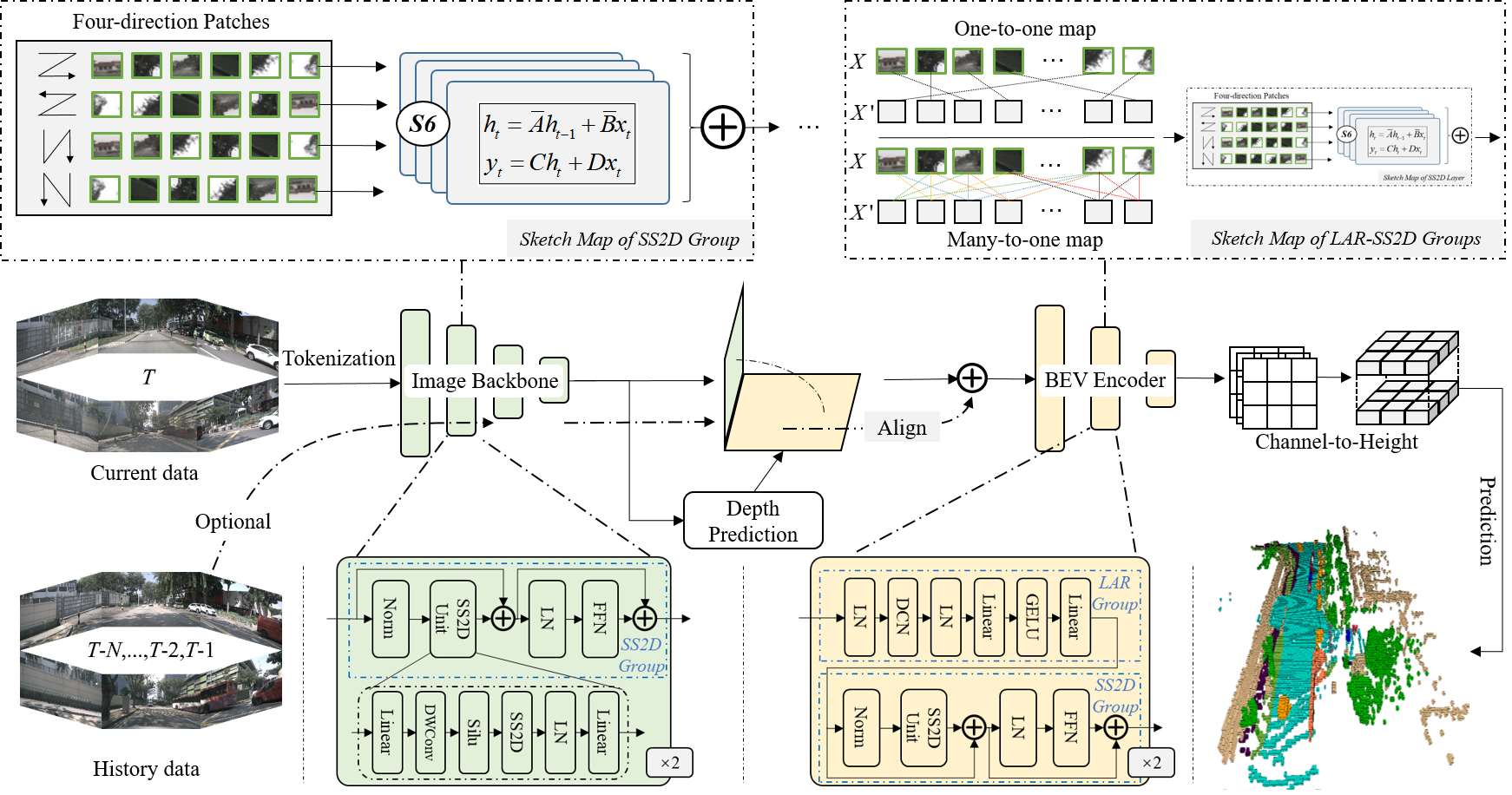}
\caption{\textcolor{black}{The framework of MambaOcc. We present the network overview in a schematic form at the top of the figure, while the bottom part of the figure provides a structural diagram of the network.}}
\label{framework}
\end{figure*}
In this section, we will elaborate on the proposed MambaOcc from four aspects: The Mamba-based image backbone (VM-Backbone) for image feature extraction, view transformation and temporal fusion module to obtain BEV-format features and aggregate multi-frame features, LAR-SS2D hybrid BEV encoder with the adaptive local re-ordering module, and the occupancy prediction head. 

\subsection{VM-Backbone}
To extract deep features from multi-view images, we utilize VMamba \cite{liu2024vmambavisualstatespace} as the backbone. Compared with convolutional networks and Transformers, Mamba-based architecture can capture contextual information in an efficient way. The core of the Mamba framework is the selective state space model (denoted as S6 following literature \cite{gu2023mamba}) which defines the update rules for the hidden states and outputs through linear time-varying systems and can be represented with four parameters $\Delta, A, B, C$ as follows:
\begin{align}
	\begin{split}
		h_t & =\overline{\boldsymbol{A}} h_{t-1}+\overline{\boldsymbol{B}} x_t, \\
            y_t & =\boldsymbol{C} h_t,\\
	\end{split}
	\label{eq_group}
\end{align}
where $x_{t}$, $h_{t}$ ad $y_{t}$ is input, hidden state, and output at timestep $t$. $\bar{A}$, $\bar{B}$ is the discretized format of $A$ and $B$:
\begin{align}
	\begin{split}
		\bar{A}&=\exp (\Delta A), \\ 
            \bar{B}&=(\Delta A)^{-1}(\exp (\Delta A)-I) \cdot \Delta B.
	\end{split}
	\label{eq_group}
\end{align}
In S6, the value $A$ is defined with model parameters and the value of $B$, $C$, and $\Delta$ are generated by a projection layer conditioned on the input $x$:
\begin{align}
	\begin{split}
		B&=s_{B}(x), \\ 
            C&=s_{C}(x), \\
            \Delta & = \tau_{\Delta}(W_{\Delta}+s_{\Delta}(x)),
	\end{split}
	\label{eq_group}
\end{align}
where $W_{\Delta}$ is system parameters and $s_{B}$, $s_{C}$ and $s_{\Delta}$ are projection layers and $\tau_{\Delta}$ is the softplus function.

S6 forms a critical component of the SS2D module (state space model for 2D data) along with the tokenization operations. To obtain the input token sequences, the multi-view images are divided into patches from four different directions following \cite{liu2024vmambavisualstatespace}. The sequences are fed into the S6 blocks individually and outputs of the S6 blocks are aligned spatially by remapping tokens to 2D feature maps. These 2D features are then summed up together to fuse different contexts. 

\subsection{View Transformation and Temporal Fusion}
In MambaOcc, we adopt the LSS \cite{huang2021bevdet} for view transformation from the image plane to the BEV plane. Firstly, the output features from the image backbone are organized into maps in 2D format. Then, depth prediction is conducted to generate a series of discrete depths for each pixel. Finally, the voxel pooling is used to aggregate the depth prediction within each grid on the predefined BEV plane. 

View transformation provides a convenient way to merge image features not only from different views but also from different timestamps if temporal fusion is activated. With the BEV features from previous frames, a feature shift operation is first conducted based on the ego movement information. Then, sampling and interpolation are applied to generate the aligned features with the current frame BEV feature maps. Finally, the aligned features are concatenated together to fuse temporal contexts. 

\subsection{LAR-SS2D Hybrid BEV Encoder}

In terms of the BEV feature extraction, we first designed the Mamba-based architecture comprised of three blocks where each block containing two SS2D groups. Considering that the SSM layer is sensitive to the order of the tokens in the sequence \cite{zhang2024voxel,wang2024pointramba}, we further explore the local adaptive pseudo-reordering (LAR) mechanism to optimize the embedding of context information. The LAR group is then used for replacing one of the SS2D groups in each block as shown in Figure \ref{framework}.

Given the input sequence $X=\{x_{0}, \dots, x_{n}\}$, if we define an index function $\pi: \{0, 1, \dots, n\} \rightarrow \{0, 1, \dots, n\}$ that represents the reordering rule, then the reordered sequence can be expressed as:

\begin{align}
	\begin{split}
		X' &= \{x_{\pi(0)}, x_{\pi(1)}, \dots, x_{\pi(n)}\},
	\end{split}
	\label{eq:reordered_sequence}
\end{align}

For strict reordering mechanism, $\pi(i)$ is a bijection from $\{0, 1, \dots, n\}$ to $\{0, 1, \dots, n\}$, indicating the new position of the element originally at position $i$. Considering the local correlation of BEV features, we made three modifications to the above reordering process as follows and propose the pseudo-reordering mechanism (For brevity, we have omitted the subscripts in the notation).  

\begin{itemize}
    \item First, we set the sorting function to be a learnable model that takes the $x$ as input. In other words, the sorting result is determined by the model parameters and the features of $x$. 
    \item 
    Second, given the difficulty of directly generating global reordering results from the input, we instead introduce mapping anchors, where the generation of $x'$ is anchored by $x$. Specifically, the anchors are used to learn a relative position offset, which is then utilized to construct the permutation function $\pi$. The permutation function $\pi$ can be expressed as:
    \begin{align}
    	\pi(\text{pos}(x)) &= \text{pos}(x) + \text{offset}(x),
    \end{align}
    
    where $\text{pos}(x)$ denotes the original position of $x$ and $\text{offset}()$ denotes the function to learn relative position offset.

    \item Third, we relaxed the mapping of $X \rightarrow X'$ from a bijection to a surjection, allowing elements with different positions in the reordered sequence to originate from the same elements in the original sequence.
\end{itemize}
The above modifications establish a flexible local pseudo-reordering mechanism. What's more, the proposed reordering process can be efficiently implemented with deformable convolution operators, which ensure high computational efficiency and maintain rapid processing speeds. 

In addition to the one-to-one mapping described above, we also propose a many-to-one mapping process. This approach aggregates features from multiple positions in the original sequence and maps them to a single position in the new sequence $X'$. To integrate features from multiple positions, we employ an attention mechanism to adaptively fuse these features, allowing the model to focus on the most relevant information. To better capture positional relationships, we introduce the position embedding in LAR and SS2D groups.

\subsection{Occupancy Prediction Head}

Following the approach outlined in FlashOcc \cite{yu2023flashocc}, we also employ a channel-to-height operation to recover height information from the channel dimension of the BEV features. This process allows us to obtain a 3D occupancy feature representation at the tail end of the network. Subsequently, we use linear layers to predict the class of each position in the 3D space, providing detailed occupancy predictions across the entire 3D volume.

\begin{table*}[htbp]
\footnotesize
\small
\setlength{\tabcolsep}{0.005\linewidth}

\def\mystrut{\rule{0pt}{1.5\normalbaselineskip}}
\centering
\caption{Performance of occupancy prediction on Occ3D-nuScenes validation dataset. The symbol $\ast$ denotes that the model is initialized with a pre-trained FCOS3D backbone. ``Cons. Veh" refers to construction vehicles, while ``Dri. Sur" is short for drivable surface. $\bullet$ indicates that the backbone has been pre-trained using nuScenes segmentation data. $\dagger$ marks the results where a camera mask was employed during the training process.} 
\begin{adjustbox}{width=2.1\columnwidth,center}
\begin{tabular}{l | c c c c| c | r r r r r r r r r r r r r r r r r r}
    \toprule[1.5pt]
    Method 
    & \rotatebox{90}{Backbone}
    & \rotatebox{90}{Image size}
    & \rotatebox{90}{Flops(G)}
    & \rotatebox{90}{Params(M)}
    & \rotatebox{90}{\textbf{mIoU}$\uparrow$}  
    & \rotatebox{90}{\textbf{others}$\uparrow$} 
    & \rotatebox{90}{\textbf{barrier}$\uparrow$}
    & \rotatebox{90}{\textbf{bicycle}$\uparrow$} 
    & \rotatebox{90}{\textbf{bus}$\uparrow$} 
    & \rotatebox{90}{\textbf{car}$\uparrow$} 
    & \rotatebox{90}{\textbf{Cons. Veh}$\uparrow$} 
    & \rotatebox{90}{\textbf{motorcycle}$\uparrow$} 
    & \rotatebox{90}{\textbf{pedestrian}$\uparrow$} 
    & \rotatebox{90}{\textbf{traffic cone}$\uparrow$} 
    & \rotatebox{90}{\textbf{trailer}$\uparrow$} 
    & \rotatebox{90}{\textbf{truck}$\uparrow$} 
    & \rotatebox{90}{\textbf{Dri. Sur}$\uparrow$} 
    & \rotatebox{90}{\textbf{other flat}$\uparrow$} 
    & \rotatebox{90}{\textbf{sidewalk}$\uparrow$} 
    & \rotatebox{90}{\textbf{terrain}$\uparrow$} 
    & \rotatebox{90}{\textbf{manmade}$\uparrow$} 
    & \rotatebox{90}{\textbf{vegetation}$\uparrow$} 
    \\
    \midrule
    MonoScene~\cite{cao2022monoscene} & R101$\ast$ & 928$\times$1600 & - & - & $\textcolor{white}{0}$6.0 & 1.7 & 7.2 & 4.2 & 4.9 & 9.3 & 5.6 & 3.9 & 3.0 & 5.9 & 4.4 & 7.1 & 14.9 & 6.3 & 7.9 & 7.4 & 1.0 & 7.6 \\
    BEVDet~\cite{huang2021bevdet} & R101$\ast$ & 704$\times$256$\textcolor{white}{0}$ & - & - & 11.7 & 2.1& 15.3& 0.0& 4.2& 13.0& 1.4& 0.0& 0.4& 0.1& 6.6& 6.7& 52.7& 19.0& 26.5& 21.8& 14.5& 15.3 \\
    BEVFormer~\cite{li2022bevformer} & R101$\ast$ & 900$\times$1600 & - & - & 23.7 & 5.0& 38.8& 10.0& 34.4& 41.1& 13.2& 16.5& 18.2& 17.8& 18.7& 27.7& 49.0& 27.7& 29.1& 25.4& 15.4& 14.5 \\
    OccFormer~\cite{zhang2023occformer} & R101$\ast$ & 928$\times$1600 & - & - & 21.9 & 5.9 & 30.2 & 12.3 & 34.4 & 39.1 & 14.4 & 16.4 & 17.2 & 9.2 & 13.9 & 26.3 & 50.9 & 30.9 & 34.6 & 22.7 & 6.7 & 6.9 \\
    TPVFormer~\cite{huang2023tri} & R101$\ast$ & 928$\times$1600 & - & - & 27.8 & 7.2 & 38.9 & 13.6 & 40.7 & 45.9 & 17.2 & 19.9 & 18.8 & 14.3 & 26.6 & 34.1 & 55.6 & 35.4 & 37.5 & 30.7 & 19.4 & 16.7 \\
    CTF-Occ~\cite{tian2024occ3d} & R101$\ast$ & 928$\times$1600 & - & - & 28.5 & 8.0 & 39.3 & 20.5 & 38.2 & 42.2 & 16.9 & 24.5 & 22.7 & 21.0 & 22.9 & 31.1 & 53.3 & 33.8 & 37.9 & 33.2 & 20.7 & 18.0 \\
    RenderOcc~\cite{pan2024renderocc} & SwinB & 512$\times$1408 & - & 121.8 & 26.1 & 4.8 & 31.7 & 10.7 & 27.6 & 26.4 & 13.8 & 18.2 & 17.6 & 17.8 & 21.1 & 23.2 & 63.2 & 36.4 & 46.2 & 44.2 & 19.5 & 20.7 \\
    PanoOcc$\dagger$~\cite{wang2024panoocc} & R101$\ast$ & 432$\times$800$\textcolor{white}{0}$ & - & 68.5 & 36.6 & 8.6 & 43.7 & 21.6 & 42.5 & 49.9 & 21.3 & 25.3 & 22.9 & 20.1 & 29.7 & 37.1 & 80.9 & 40.3 & 49.6 & 52.8 & 39.8 & 35.8 \\
    PanoOcc$\dagger$~\cite{wang2024panoocc} & R101$\ast$ & 864$\times$1600 & - & 98.2 & 41.6 & 11.9 & 49.8 & 28.9 & 45.4 & 54.7 & 25.2 & 32.9 & 28.8 & 30.7 & 33.8 & 41.3 & 83.1 & 45.0 & 53.8 & 56.1 & 45.1 & 40.1 \\
    PanoOcc$\dagger$~\cite{wang2024panoocc} & R101$\bullet$ & 864$\times$1600 & - & 98.2 & 42.2 & 11.6 & 50.4 & 29.6 & 49.4 & 55.5 & 23.2 & 33.2 & 30.5 & 30.9 & 34.4 & 42.5 & 83.3 & 44.2 & 54.4 & 56.0 & 45.9 & 40.4 \\

    FlashOcc$\dagger$:M3\cite{yu2023flashocc} & SwinB                    & 512$\times$1408 &1467.5 & 137.1 & 43.3 & 12.9 & 50.5 & 27.4 & 52.4 & 55.6 & 27.4 & 29.0 & 28.6 & 29.7 & 37.5 & 43.1 & 84.0 & 46.5 & 56.3 & 59.3 & 51.0 & 44.6 \\
    \midrule
    \bf{MambaOcc$\dagger$}    & VmambaT & 512$\times$1408 & 893.8 & 79.5 & 43.4 & 12.5 & 50.5 & 26.2 & 50.6 & 55.9 & 28.6 & 27.1 & 29.1 & 27.2 & 35.7 & 43.1 & 85.1 & 47.8 & 57.8 & 61.4 & 53.3 & 46.4 \\
    \bf{MambaOcc-Large$\dagger$}    & VmambaT & 512$\times$1408 & 1002  & 119  & \textbf{44.1} & 12.8 & 50.5 & 28.1 & 50.8 & 56.4 & 29.4 & 27.9 & 29.6 & 28.1 & 36.7 & 43.8 & 85.6 & 48.9 & 58.3 & 61.8 & 53.8 & 46.6 \\

\bottomrule[1.5pt]
\end{tabular}
\end{adjustbox}
\vspace{-0.3cm}
\label{table:sota_occ_eval}
\end{table*}

\section{Experiments}

\begin{figure*}[htbp]
\centering
\includegraphics[width=7in]{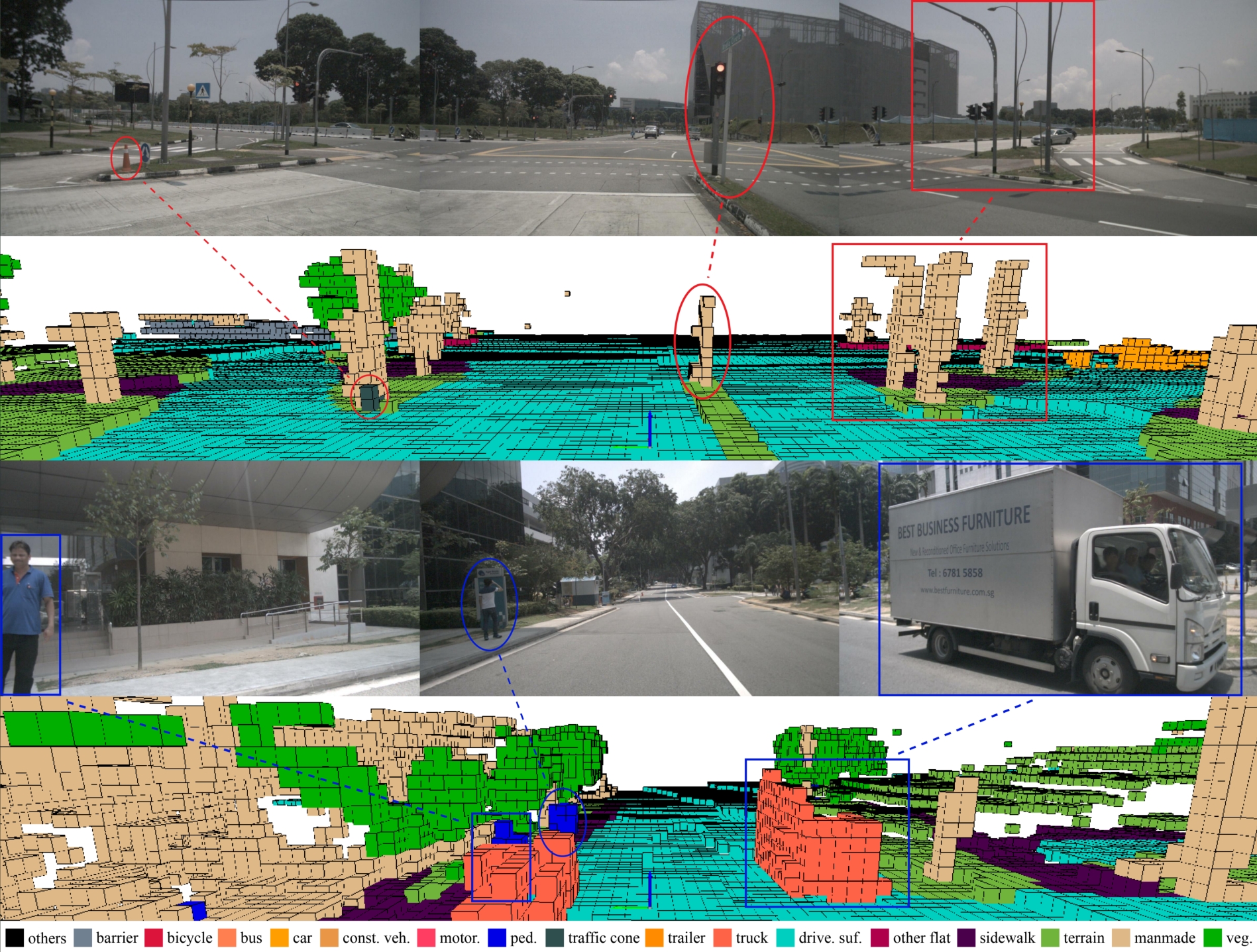}
\caption{\textcolor{black}{The demo of occupancy prediction of MambaOcc.}}
\label{demo-mamba}
\end{figure*}

\begin{figure*}[htbp]
\centering
\includegraphics[width=7in]{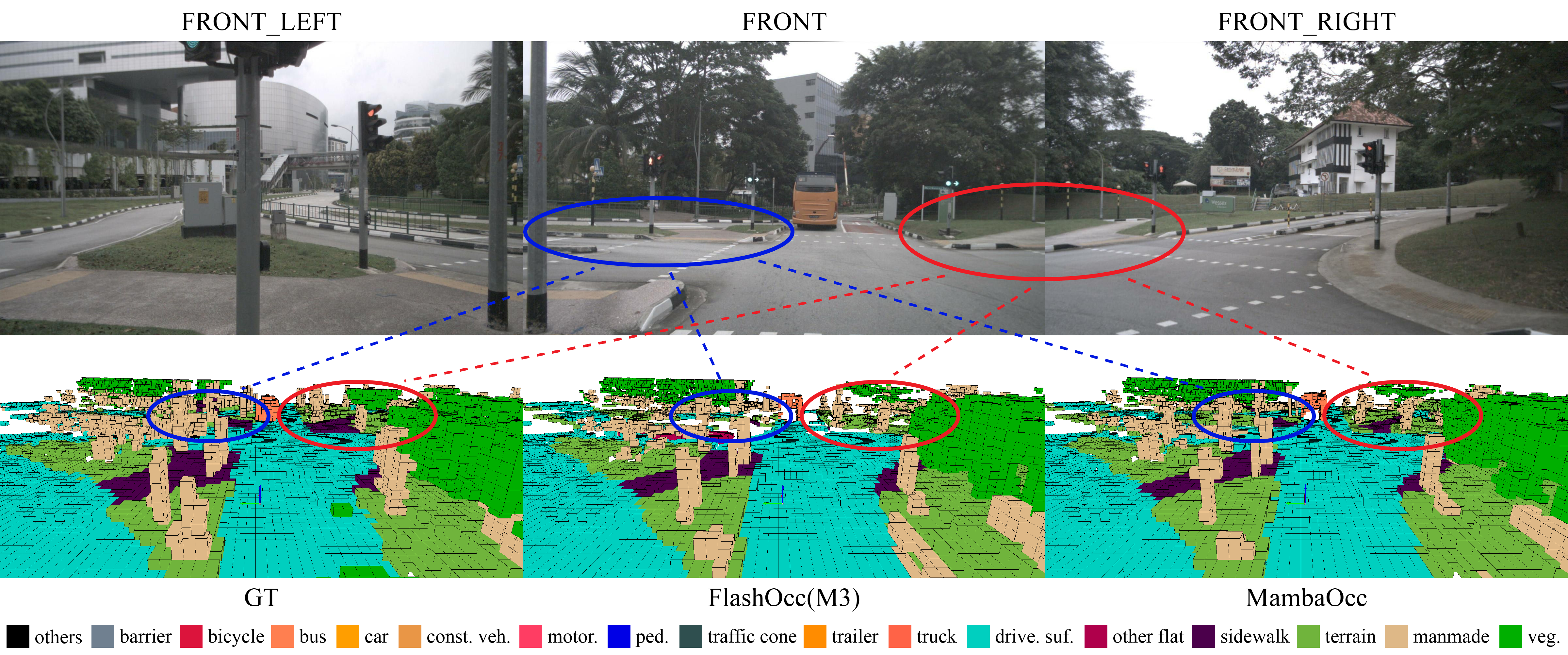}
\caption{\textcolor{black}{The visualized comparison of occupancy prediction between MambaOcc and FlashOcc.}}
\label{demo-comp}
\end{figure*}

\subsection{Dataset and Training Details}
To validate the performance of MambaOcc, We conducted experiments on the widely used Occ3D-nuScenes dataset, which consists of 700 sequences as the training set and 150 sequences as the validation set. Each driving sequence comprises 20 seconds of annotated perception data, recorded at a frequency of 2 Hz. The data collection vehicle is equipped with six cameras, offering a full 360-degree view of the surroundings. During the generation of semantic occupancy labels (covering 17 categories) of the Occ3D-nuScenes dataset, the voxel size is set at 0.4 meters for X, Y, and Z dimensions simultaneously. The valid perception range extends from -40.0m to 40.0m along both the X and Y axes of the BEV plane, and from -1.0m to 5.4m along the Z axis. For evaluation purposes, the mean Intersection over Union (mIoU) across all categories is used as the performance metric.

In the training of MambaOcc, we set the total epochs to 24 and the optimizer as AdamW for all our experiments. 8 NVIDIA V100 GPUs are utilized with a learning rate of 0.0001. For the image backbone of MambaOcc, without explicitly stating it, we adopt the Vmamba-Tiny (abbreviated as VmambaT) in our experiments considering its lightweight nature.

\subsection{Comparison with SOTA appropachs}
In the following part of this section, we mainly compare MambaOcc with other SOTA approaches (More experiments of the proposed MambaOcc networks with other settings will be elaborated on in the next section of ablation studies). The image backbone is pre-trained on the ImageNet dataset to provide better initialization of the training process. The number of channels in each BEV block is set as [128, 256, 512] for the basic MambaOcc and set as [256, 512, 1024] (the corresponding SS2D group is denoted as SS2D-L) for MambaOcc-Large. For the DCN layer in LAR Group, we set the kernel size to $3 \time 3$ by default. 

From the results shown in Table \ref{table:sota_occ_eval}, it is evident that the MambaOcc method offers significant advantages in terms of computational efficiency and parameter count compared to the state-of-the-art methods. Compared to FlashOcc with Swin Transformer as the backbone in Table \ref{table:sota_occ_eval}), MambaOcc achieves better performance while reducing parameters by 42\% and computational cost by 39\%. Additionally, MambaOcc-Large surpasses FlashOcc by 0.77 mIoU, reducing parameters by 14\% and computational cost by 32\%. Compared to PanoOcc with ResNet-101 as the backbone, MambaOcc outperforms it by 1.23 mIoU while reducing parameters by 19\%. These results show that the proposed Mamba framework offers significant advantages in terms of parameter amount, computational efficiency, and perceptual capability compared to CNN and Transformer-based methods.

Figure \ref{demo-mamba} showcases the occupancy predictions generated by the proposed MambaOcc model. As illustrated, MambaOcc delivers precise perception results for typical objects like humans and vehicles, while also effectively detecting irregularly structured objects such as utility poles, traffic lights, and road cones. In Figure \ref{demo-comp}, a comparison of occupancy prediction results between MambaOcc and FlashOcc is presented. The comparison highlights MambaOcc's superior performance in long-range planar perception, offering more comprehensive ground predictions, whereas FlashOcc often predicts these regions as empty.

\section{Ablation Studies}
In this part, we analyze the effectiveness of different modules of MambaOcc and conduct several studies on different settings of each module. Without explicitly stating it, we remove the temporal fusion module and use an input size of $[256\times704]$ during the ablation studies to accelerate the process of experiments.

\begin{table}[th]
\centering
\caption{Ablation study for each component on the Occ3D-NuScenes.}
\begin{tabular}{ccc|c}
\hline
\textbf{VM-Backbone} & \textbf{LAR-SS2D BEV}  & \textbf{PE}  &  \textbf{mIoU ↑}\\ \hline
 &  &  & 30.20  \\ 
  & \checkmark & \checkmark & 31.52 \\
\checkmark &  &  & 34.16  \\ 
\checkmark & \checkmark &  & 35.28 \\

\checkmark & \checkmark & \checkmark & 35.41 \\ \hline
\end{tabular}
\label{table_ablation}
\end{table}
\subsection{Effectiveness of the Proposed Modules}

To clearly demonstrate the contribution of each component within MambaOcc, we present ablation studies in Table \ref{table_ablation} to underscore the effectiveness of each module. In these studies, the FlashOcc model with a ResNet-50 image backbone is used as the baseline. Notably, replacing the CNN with Mamba results in a significant mIoU increase of 3.96, highlighting the effectiveness of the Mamba architecture. Additionally, the proposed LAR-SS2D BEV encoder contributes an extra 1.12 mIoU gain over the CNN-based encoder. Moreover, incorporating positional encoding (denoted as PE in Table \ref{table_ablation}) provides a further modest enhancement in performance.

\subsection{Initialization of Image Backbone.} Table \ref{img_init} presents the impact of different initialization methods for the image backbone on occupancy prediction performance. For clarity, CNN is used as the BEV encoder in all the experiments discussed in this subsection. It is evident that a good parameter initialization method significantly influences the performance. Initializing the occupancy prediction network with pre-training on ImageNet classification yields substantially better results compared with random initialization for both mamba and convolution networks. For example, MambaOcc with ImageNet-pretrained VM-Backbone gets 10.01 higher performance in terms of mIoU compared with that initialized with random values. 

\begin{table}[!ht]
    \centering

    \caption{Ablation study on the initialization method of image backbone.}
    \begin{tabular}{cc|c}
    \hline
        \textbf{Image-Backbone} & \textbf{Initialization}   &  \textbf{mIoU↑} \\ \hline
        ResNet50 & Random  & 22.58 \\ 
        ResNet50 & ImageNet   & 30.20 \\ 
        \hline
        VMambaT & Random & 24.15 \\ 
        VMambaT & ImageNet   & 34.16 \\
        \hline
    \end{tabular}
    \label{img_init}
\end{table}

\subsection{Influence of Different BEV Encoders} 
Table \ref{bev_encoder} shows the impact of different BEV encoders on occupancy prediction (Each block in the BEV encoder is designed with two groups.). For clarity, the VMamba-T network was used as the image backbone in all experiments discussed in this subsection. We compared the results of using a pure CNN and a pure SS2D as the BEV encoder, along with the outcomes of a hybrid structure combining CNN and SS2D, and the proposed combination of LAR and SS2D. As shown in the table, the structure of the BEV encoder significantly impacts occupancy prediction performance. The pure SS2D outperforms the pure CNN, with a 0.56 mIoU improvement. The hybrid CNN-SS2D architecture performs better than both the pure CNN and the pure SS2D, with mIoU improvements of 0.77 and 0.21, respectively. The proposed LAR-SS2D hybrid architecture achieves the best results, surpassing the CNN-SS2D hybrid by 0.48 mIoU.

\begin{table}[!ht]
    \centering
    \caption{Ablation study on the influence of different BEV encoders.}
    \begin{tabular}{cc|c}
    \hline
    \textbf{Image-Backbone}  & \textbf{BEV-Encoder} &  \textbf{mIoU↑}\\ \hline
        VMambaT & CNN-CNN&  34.16 \\ 
        VMambaT &  SS2D-SS2D  & 34.72 \\ 
        VMambaT & CNN-SS2D &  34.93 \\ 
        VMambaT & LAR-SS2D & 35.41 \\ 
        \hline
    \end{tabular}
     \label{bev_encoder}
\end{table}

\subsection{Influence of the Model Scale}
Table \ref{scale} presents the comparative results of models with different scales. We selected various mamba-based image backbones and BEV encoders with different sizes for comparison. For LAR-SS2D-L, we double the number of channels in the BEV encoder compared to the standard LAR-SS2D (It means that the numbers of channels in the BEV block are [256, 512, 1024] for LAR-SS2D-L, instead of [128, 256, 512]). As can be observed, the occupancy prediction performance improves as the model scale increases, though the mIoU gains are not particularly significant. Given that the parameter count and computational cost also increase with model size, selecting the VMambaT as the backbone is a more cost-effective option when training resources are limited.
\begin{table}[!ht]
    \centering
    
    \caption{Performance and computation cost comparison with MambaOcc with different model scales.}
    \begin{tabular}{c|c|cc|c}
    \hline
        \textbf{Image} & \textbf{BEV} & \textbf{FLOPs} & \textbf{Params} &  \multirow{2}{*}{\textbf{mIoU↑}} \\
        \textbf{Backbone} & \textbf{Encoder} & (G) & (M)& \\ \hline
    
        VMambaT  & LAR-SS2D & 262 & 52 & 35.41\\ 
        VMambaT  & LAR-SS2D-L & 339 & 90 & 35.80\\ 
        VMambaS  & LAR-SS2D & 345 & 71 & 35.87 \\ 
        VMambaB  & LAR-SS2D & 489 & 109 & 36.09 \\ 
        \hline
    \end{tabular}
    \label{scale}
\end{table}

\subsection{Mapping Mechanism in LAR Layer}
In this subsection, we compare the effects of different mapping methods within the LAR layer. For the many-to-one mapping, we conduct experiments with different entry numbers of 3$\times$ 3 and 5$\times$5, where information from multiple positions in the original sequence is weighted and fused before being mapped to the same position in the new sequence. The results in Table \ref{mapping} show that many-to-one mapping methods outperform the one-to-one approach. Specifically, the 5$\times$5 and 3$\times$3 configurations improve performance by 0.07 and 0.32 mIoU, respectively, compared to the one-to-one method. These findings suggest that many-to-one mapping can be an effective strategy for enhancing performance.
\begin{table}[h]
\centering
\caption{Effects of different mapping mechanisms in LAR layer.}
\begin{tabular}{cc|c}
\hline
\textbf{Type} & \textbf{Entry number} &  \textbf{mIoU ↑}\\ \hline

One-to-one & 1 & 35.34  \\ 
Many-to-one & 3 $\times$ 3 & 35.41 \\ 
Many-to-one & 5 $\times$ 5 & 35.66 \\
  \hline
\end{tabular}
\label{mapping}
\end{table}

\subsection{Number of sequence directions in SS2D group}
This section presents a comparison between using four-directional sequences and single-directional sequences (the first direction shown in Figure \ref{framework}) in the SS2D group. The results in Table \ref{directions} show that when using a single-directional sequence, the introduction of LAR outperforms both the purely single-directional SS2D and the combination of CNN with single-directional SS2D. Moreover, our proposed LAR combined with the single-directional sequence even surpasses the four-directional SS2D approach.

\begin{table}[!ht]
    \centering
    \caption{Performance comparison of the SS2D layer with different numbers of directions. The symbol $\ast$ indicates that the SS2D layer is one-directional.}
    \begin{tabular}{c|c}
    \hline
         \textbf{BEV Encoder} &   {\textbf{mIoU↑}} \\ 
         \hline
        CNN-CNN & 34.16 \\
        SS2D-SS2D & 34.72\\
        \hline
        SS2D$\ast$-SS2D$\ast$ & 34.45\\
        CNN-SS2D$\ast$ & 34.70 \\
        LAR-SS2D$\ast$ & 35.16\\ 
        \hline
    \end{tabular}
    \label{directions}
\end{table}

\subsection{Effectiveness of Temporal Fusion Module}
Table \ref{temporal} presents a comparison of the results before and after incorporating the temporal fusion module (MambaOcc-4D denotes the model with temporal fusion). The introduction of historical information significantly enhances the performance of the occupancy prediction, resulting in a 4.7-point increase in mIoU.

\begin{table}[!ht]
    \centering

    \caption{The results before and after incorporating the temporal fusion module.}
    \begin{tabular}{c|c}
    \hline
        \textbf{Method} &  \textbf{mIoU↑} \\ \hline
        MambaOcc &35.41 \\ 
        MambaOcc-4D &39.78 \\ \hline
    \end{tabular}
    \label{temporal}
\end{table}

\subsection{Visualization of Mapping Mechanisms in LAR Layer}
\begin{figure}[htbp]
\centering
\includegraphics[width=3.5in]{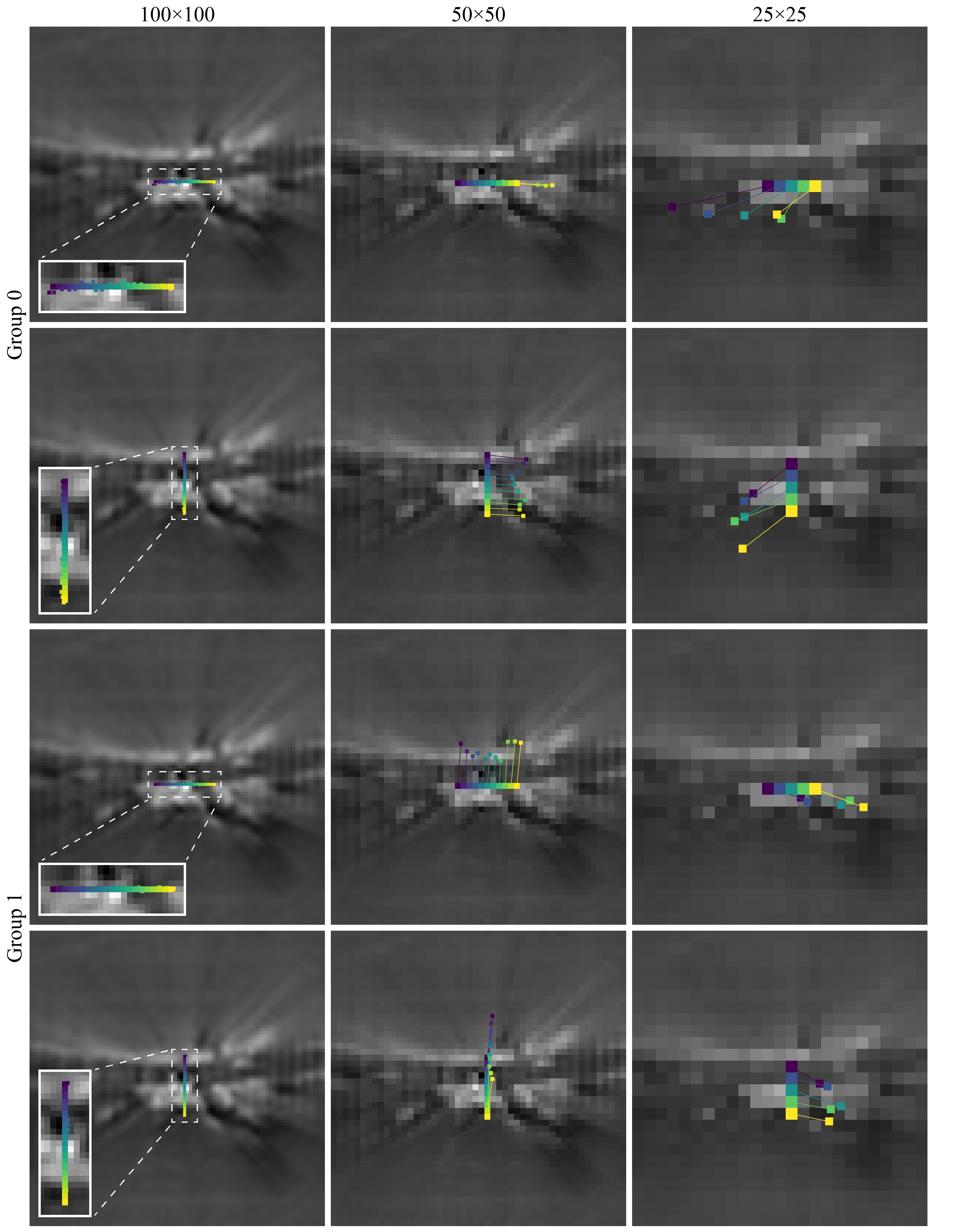}
\caption{\textcolor{black}{The visualization of one-to-one mapping (group 0 and group 1).}}
\label{vis_dcn1}
\end{figure}

\begin{figure}[htbp]
\centering
\includegraphics[width=3.5in]{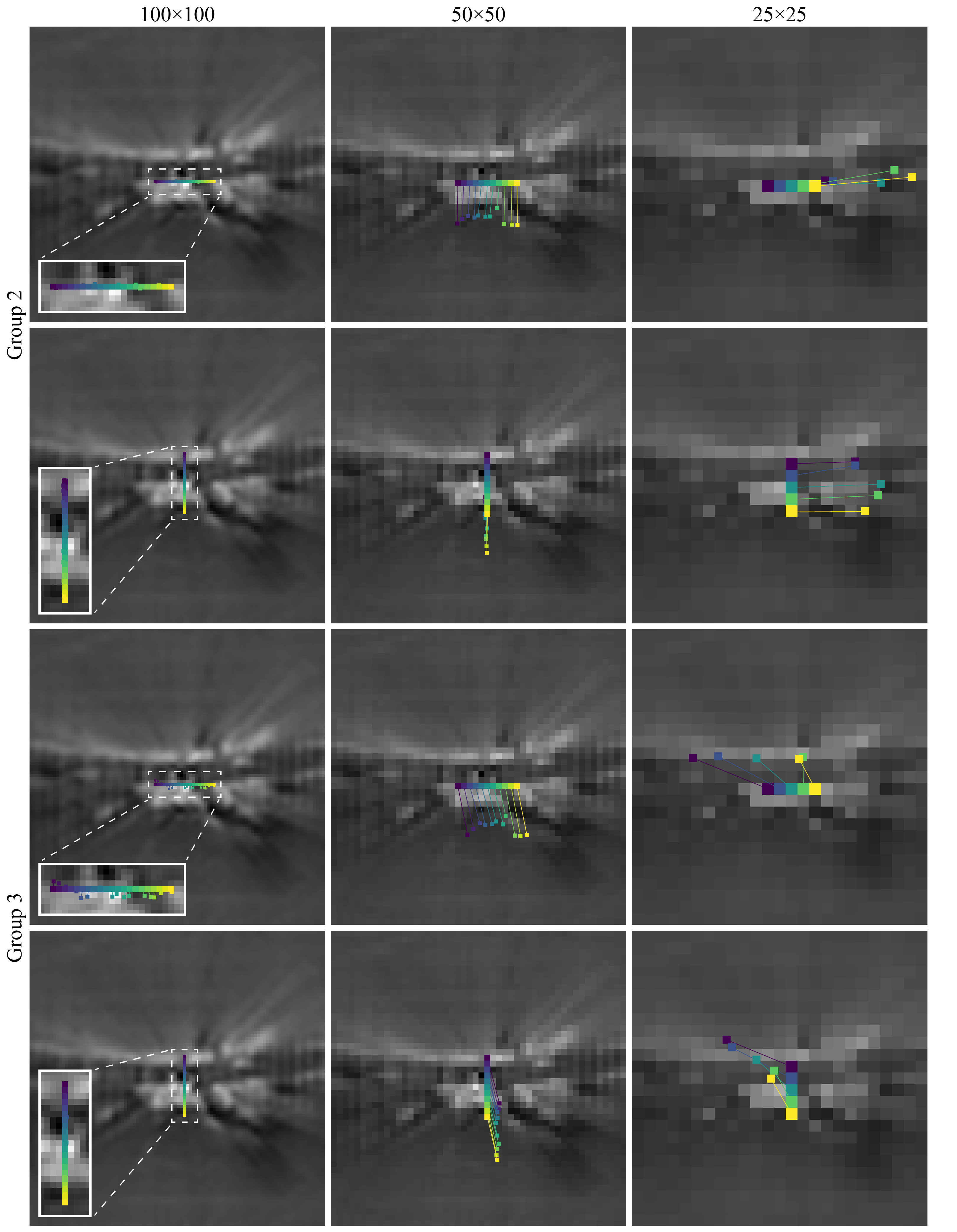}
\caption{\textcolor{black}{The visualization of one-to-one mapping (group 2 and group 3).}}
\label{vis_dcn2}
\end{figure}

In this section, we analyze the mapping function within the LAR layer by visualizing the positional correspondence between $X$ and $X'$. To gain a more comprehensive understanding of the mapping patterns, we applied four different mapping modes to each LAR layer and executed these modes across grouped feature channels. The results are visualized in Figures \ref{vis_dcn1} and \ref{vis_dcn2} (Figure \ref{vis_dcn1} corresponds to groups 0 and 1, and Figure \ref{vis_dcn2} corresponds to groups 2 and 3). In the figures, we use larger, continuous horizontal or vertical color blocks to represent the positions in the updated sequence $X'$, while smaller, discrete color blocks indicate the positions in the original sequence $X$. As we can observe, the LAR layer attempts to extract relevant information from different spatial locations and map it to the current position. Moreover, this mapping pattern varies depending on the depth and resolution of the feature maps. On lower-level feature maps, the mapping function tends to favor identity mapping, while on deeper feature maps, the mapping process becomes more aggressive, establishing connections between more distant points. We also observed that the mapping patterns vary significantly across different groups, which suggests that this diversity might help the model establish more comprehensive connections between elements.

\section{Conclusion}
In this paper, we propose the first Mamba-based occupancy prediction network, named MambaOcc. Thanks to Mamba's linear attention mechanism and long-range context modeling approach, MambaOcc surpasses CNN-based methods while offering higher parameter efficiency compared to Transformer-based approaches. Additionally, to address the sensitivity of the Mamba model to sequence order, we introduce a local adaptive reordering method. This method improves the model's performance by reconstructing sequence order from context through an input-dependent learning mechanism.

\ifCLASSOPTIONcaptionsoff
  \newpage
\fi



%
\bibliographystyle{IEEEtran}
\bibliography{mambaocc}

\begin{thebibliography}{10}
\providecommand{\url}[1]{#1}
\csname url@samestyle\endcsname
\providecommand{\newblock}{\relax}
\providecommand{\bibinfo}[2]{#2}
\providecommand{\BIBentrySTDinterwordspacing}{\spaceskip=0pt\relax}
\providecommand{\BIBentryALTinterwordstretchfactor}{4}
\providecommand{\BIBentryALTinterwordspacing}{\spaceskip=\fontdimen2\font plus
\BIBentryALTinterwordstretchfactor\fontdimen3\font minus \fontdimen4\font\relax}
\providecommand{\BIBforeignlanguage}[2]{{%
\expandafter\ifx\csname l@#1\endcsname\relax
\typeout{** WARNING: IEEEtran.bst: No hyphenation pattern has been}%
\typeout{** loaded for the language `#1'. Using the pattern for}%
\typeout{** the default language instead.}%
\else
\language=\csname l@#1\endcsname
\fi
#2}}
\providecommand{\BIBdecl}{\relax}
\BIBdecl

\bibitem{huang2023tri}
Y.~Huang, W.~Zheng, Y.~Zhang, J.~Zhou, and J.~Lu, ``Tri-perspective view for vision-based 3d semantic occupancy prediction,'' in \emph{Proceedings of the IEEE/CVF conference on computer vision and pattern recognition}, 2023, pp. 9223--9232.

\bibitem{ming2024inversematrixvt3d}
Z.~Ming, J.~S. Berrio, M.~Shan, and S.~Worrall, ``Inversematrixvt3d: An efficient projection matrix-based approach for 3d occupancy prediction,'' \emph{arXiv preprint arXiv:2401.12422}, 2024.

\bibitem{yu2023flashocc}
Z.~Yu, C.~Shu, J.~Deng, K.~Lu, Z.~Liu, J.~Yu, D.~Yang, H.~Li, and Y.~Chen, ``Flashocc: Fast and memory-efficient occupancy prediction via channel-to-height plugin,'' \emph{arXiv preprint arXiv:2311.12058}, 2023.

\bibitem{liu2024swin}
J.~Liu, H.~Yang, H.-Y. Zhou, Y.~Xi, L.~Yu, Y.~Yu, Y.~Liang, G.~Shi, S.~Zhang, H.~Zheng \emph{et~al.}, ``Swin-umamba: Mamba-based unet with imagenet-based pretraining,'' \emph{arXiv preprint arXiv:2402.03302}, 2024.

\bibitem{ma2024cotr}
Q.~Ma, X.~Tan, Y.~Qu, L.~Ma, Z.~Zhang, and Y.~Xie, ``Cotr: Compact occupancy transformer for vision-based 3d occupancy prediction,'' in \emph{Proceedings of the IEEE/CVF Conference on Computer Vision and Pattern Recognition}, 2024, pp. 19\,936--19\,945.

\bibitem{gu2023mamba}
A.~Gu and T.~Dao, ``Mamba: Linear-time sequence modeling with selective state spaces,'' \emph{arXiv preprint arXiv:2312.00752}, 2023.

\bibitem{liu2024vmambavisualstatespace}
Y.~Liu, Y.~Tian, Y.~Zhao, H.~Yu, L.~Xie, Y.~Wang, Q.~Ye, and Y.~Liu, ``Vmamba: Visual state space model,'' \emph{arXiv preprint arXiv:2401.10166}, 2024.

\bibitem{cao2022monoscene}
A.-Q. Cao and R.~De~Charette, ``Monoscene: Monocular 3d semantic scene completion,'' in \emph{Proceedings of the IEEE/CVF Conference on Computer Vision and Pattern Recognition}, 2022, pp. 3991--4001.

\bibitem{zhang2023occformer}
Y.~Zhang, Z.~Zhu, and D.~Du, ``Occformer: Dual-path transformer for vision-based 3d semantic occupancy prediction,'' in \emph{Proceedings of the IEEE/CVF International Conference on Computer Vision}, 2023, pp. 9433--9443.

\bibitem{wang2023openoccupancy}
X.~Wang, Z.~Zhu, W.~Xu, Y.~Zhang, Y.~Wei, X.~Chi, Y.~Ye, D.~Du, J.~Lu, and X.~Wang, ``Openoccupancy: A large scale benchmark for surrounding semantic occupancy perception,'' in \emph{Proceedings of the IEEE/CVF International Conference on Computer Vision}, 2023, pp. 17\,850--17\,859.

\bibitem{tian2024occ3d}
X.~Tian, T.~Jiang, L.~Yun, Y.~Mao, H.~Yang, Y.~Wang, Y.~Wang, and H.~Zhao, ``Occ3d: A large-scale 3d occupancy prediction benchmark for autonomous driving,'' \emph{Advances in Neural Information Processing Systems}, vol.~36, 2024.

\bibitem{wei2023surroundocc}
Y.~Wei, L.~Zhao, W.~Zheng, Z.~Zhu, J.~Zhou, and J.~Lu, ``Surroundocc: Multi-camera 3d occupancy prediction for autonomous driving,'' in \emph{Proceedings of the IEEE/CVF International Conference on Computer Vision}, 2023, pp. 21\,729--21\,740.

\bibitem{gu2021efficiently}
A.~Gu, K.~Goel, and C.~R{\'e}, ``Efficiently modeling long sequences with structured state spaces,'' \emph{arXiv preprint arXiv:2111.00396}, 2021.

\bibitem{gu2021combining}
A.~Gu, I.~Johnson, K.~Goel, K.~Saab, T.~Dao, A.~Rudra, and C.~R{\'e}, ``Combining recurrent, convolutional, and continuous-time models with linear state space layers,'' \emph{Advances in neural information processing systems}, vol.~34, pp. 572--585, 2021.

\bibitem{gupta2022diagonal}
A.~Gupta, A.~Gu, and J.~Berant, ``Diagonal state spaces are as effective as structured state spaces,'' \emph{Advances in Neural Information Processing Systems}, vol.~35, pp. 22\,982--22\,994, 2022.

\bibitem{li2022makes}
Y.~Li, T.~Cai, Y.~Zhang, D.~Chen, and D.~Dey, ``What makes convolutional models great on long sequence modeling?'' \emph{arXiv preprint arXiv:2210.09298}, 2022.

\bibitem{orvieto2023resurrecting}
A.~Orvieto, S.~L. Smith, A.~Gu, A.~Fernando, C.~Gulcehre, R.~Pascanu, and S.~De, ``Resurrecting recurrent neural networks for long sequences,'' in \emph{International Conference on Machine Learning}.\hskip 1em plus 0.5em minus 0.4em\relax PMLR, 2023, pp. 26\,670--26\,698.

\bibitem{smith2022simplified}
J.~T. Smith, A.~Warrington, and S.~W. Linderman, ``Simplified state space layers for sequence modeling,'' \emph{arXiv preprint arXiv:2208.04933}, 2022.

\bibitem{nguyen2022s4nd}
E.~Nguyen, K.~Goel, A.~Gu, G.~Downs, P.~Shah, T.~Dao, S.~Baccus, and C.~R{\'e}, ``S4nd: Modeling images and videos as multidimensional signals with state spaces,'' \emph{Advances in neural information processing systems}, vol.~35, pp. 2846--2861, 2022.

\bibitem{zhu2024vision}
L.~Zhu, B.~Liao, Q.~Zhang, X.~Wang, W.~Liu, and X.~Wang, ``Vision mamba: Efficient visual representation learning with bidirectional state space model,'' \emph{arXiv preprint arXiv:2401.09417}, 2024.

\bibitem{ma2024u}
J.~Ma, F.~Li, and B.~Wang, ``U-mamba: Enhancing long-range dependency for biomedical image segmentation,'' \emph{arXiv preprint arXiv:2401.04722}, 2024.

\bibitem{huang2021bevdet}
J.~Huang, G.~Huang, Z.~Zhu, Y.~Ye, and D.~Du, ``Bevdet: High-performance multi-camera 3d object detection in bird-eye-view,'' \emph{arXiv preprint arXiv:2112.11790}, 2021.

\bibitem{zhang2024voxel}
G.~Zhang, L.~Fan, C.~He, Z.~Lei, Z.~Zhang, and L.~Zhang, ``Voxel mamba: Group-free state space models for point cloud based 3d object detection,'' \emph{arXiv preprint arXiv:2406.10700}, 2024.

\bibitem{wang2024pointramba}
Z.~Wang, Z.~Chen, Y.~Wu, Z.~Zhao, L.~Zhou, and D.~Xu, ``Pointramba: A hybrid transformer-mamba framework for point cloud analysis,'' \emph{arXiv preprint arXiv:2405.15463}, 2024.

\bibitem{li2022bevformer}
Z.~Li, W.~Wang, H.~Li, E.~Xie, C.~Sima, T.~Lu, Y.~Qiao, and J.~Dai, ``Bevformer: Learning bird’s-eye-view representation from multi-camera images via spatiotemporal transformers,'' in \emph{European conference on computer vision}.\hskip 1em plus 0.5em minus 0.4em\relax Springer, 2022, pp. 1--18.

\bibitem{pan2024renderocc}
M.~Pan, J.~Liu, R.~Zhang, P.~Huang, X.~Li, H.~Xie, B.~Wang, L.~Liu, and S.~Zhang, ``Renderocc: Vision-centric 3d occupancy prediction with 2d rendering supervision,'' in \emph{2024 IEEE International Conference on Robotics and Automation (ICRA)}.\hskip 1em plus 0.5em minus 0.4em\relax IEEE, 2024, pp. 12\,404--12\,411.

\bibitem{wang2024panoocc}
Y.~Wang, Y.~Chen, X.~Liao, L.~Fan, and Z.~Zhang, ``Panoocc: Unified occupancy representation for camera-based 3d panoptic segmentation,'' in \emph{Proceedings of the IEEE/CVF conference on computer vision and pattern recognition}, 2024, pp. 17\,158--17\,168.

\end{thebibliography}



%








\end{document}